\PassOptionsToPackage{prologue,dvipsnames}{xcolor}
\documentclass[conference]{IEEEtran}
\IEEEoverridecommandlockouts

\usepackage{cite}
\usepackage{amsmath,amssymb,amsfonts}
\usepackage{algorithmic}
\usepackage{booktabs}
\usepackage{graphicx}
\usepackage{textcomp}
\usepackage{xcolor}

\def\BibTeX{{\rm B\kern-.05em{\sc i\kern-.025em b}\kern-.08em
    T\kern-.1667em\lower.7ex\hbox{E}\kern-.125emX}}

\usepackage{caption}
\usepackage{tablefootnote}

\makeatletter
\patchcmd{\@makecaption}
  {\scshape}
  {}
  {}
  {}
\makeatother

\newlength\myheight
\newlength\mydepth
\settototalheight\myheight{Xygp}
\newcommand{\red}[1]{\textcolor{black}{#1}}
\definecolor{blue}{RGB}{91,186,240}
\newcommand{\green}[1]{\textcolor{ForestGreen}{#1}}
\newcommand{\orange}[1]{\textcolor{orange}{#1}}
    
\begin{document}

\title{Pseudo Dataset Generation for Out-of-domain Multi-Camera View Recommendation}

\author{\IEEEauthorblockN{Kuan-Ying Lee\textsuperscript{1}, Qian Zhou\textsuperscript{2}, Klara Nahrstedt\textsuperscript{1}}
\IEEEauthorblockA{
\textit{\textsuperscript{1}University of Illinois, Urbana-Champaign}\\
\textit{\textsuperscript{2}City University of Hong Kong}\\
kylee5@illinois.edu, qiazhou@cityu.edu.hk, klara@illinois.edu
}}

\maketitle

\begin{abstract}
Multi-camera systems are indispensable in movies, TV shows, and other media.
Selecting the appropriate camera at every timestamp has a decisive impact on production quality and audience preferences. Learning-based view recommendation frameworks can assist professionals in decision-making.
However, they often struggle outside of their training domains. The scarcity of labeled multi-camera view recommendation datasets exacerbates the issue.
Based on the insight that many videos are edited from the original multi-camera videos, we propose transforming regular videos into pseudo-labeled multi-camera view recommendation datasets.
Promisingly, by training the model on pseudo-labeled datasets stemming from videos in the target domain, we achieve a 68\% relative improvement in the model's accuracy in the target domain and bridge the accuracy gap between in-domain and never-before-seen domains.

\end{abstract}


\section{Introduction} \label{sec:intro}
Multi-camera systems capturing the same scene provide different viewing perspectives and play significant roles in movies, broadcasts, news shows, etc. \cite{stoll2023automatic}.
However, to benefit from multi-camera systems, expertise from cinematography professionals is heavily required. For instance, a video editor has to spend several hours watching pre-recorded videos from all cameras, determining which portions to use, and editing a single-track movie that best narrates the story. 

% In this work, we present a 
Learning-based multi-camera view recommendation could assist professionals in choosing which camera to switch to by analyzing the past frames that have been selected, thus improving their efficiency.
To facilitate the development of learning-based multi-camera recommendation, Rao et al. propose a dataset called TV shows Multi-camera Editing Dataset (TVMCE) \cite{tvmce}, providing subsampled frames of multi-camera videos and the camera transitions determined by professional videographers.
\begin{figure}
    \centerline{
\includegraphics[width=\linewidth]{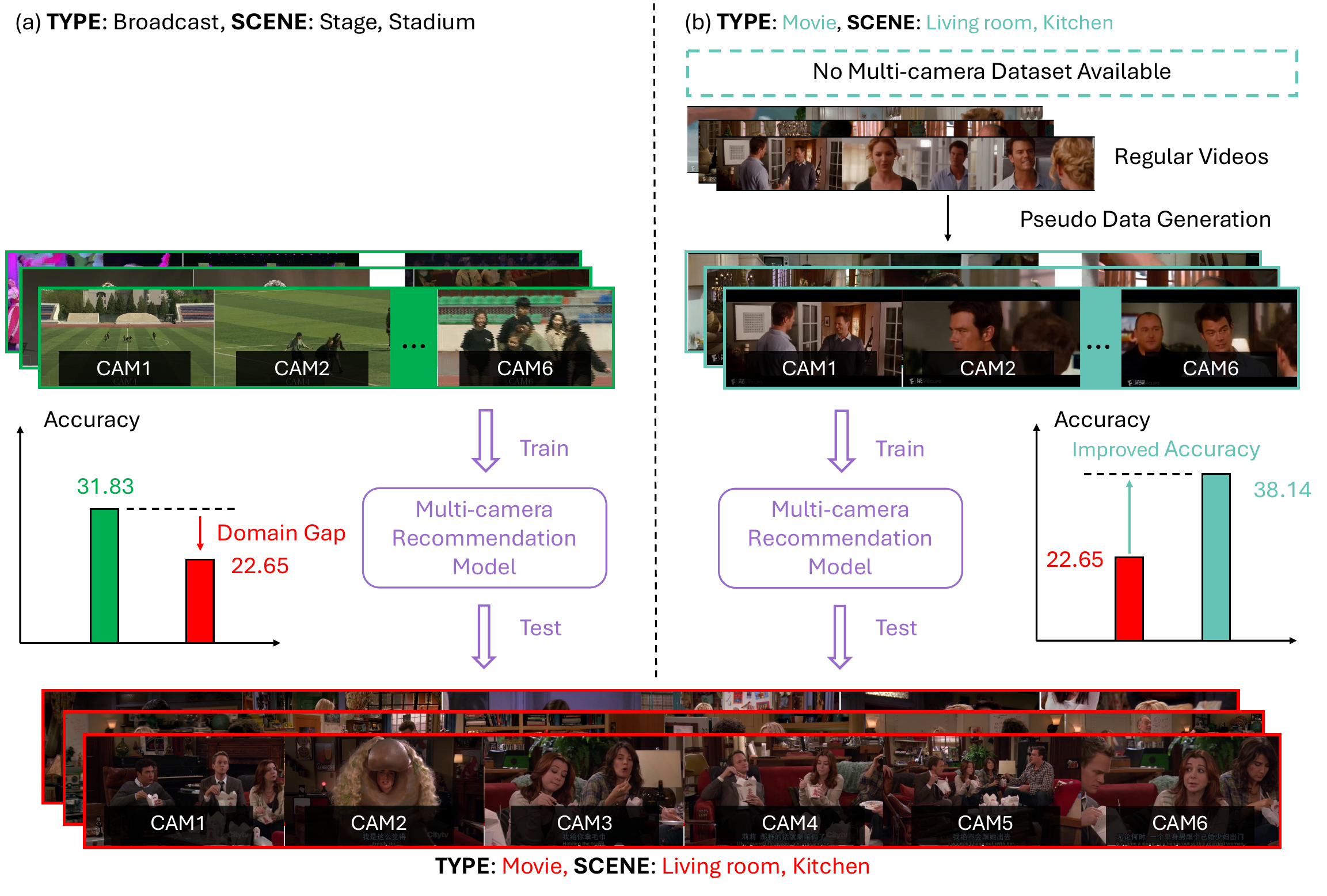}}
    \caption{
    \textbf{} (a) A model trained on a labeled multi-camera editing dataset of a particular domain generalizes poorly to a never-before-seen domain and the accuracy drops significantly.
    (b) Our proposed method leverages regular videos to generate pseudo-labeled datasets for the target domain and improve the model's accuracy. \textbf{[Best viewed in color.]}
    }
    \label{fig:teaser}
\end{figure}

However, videos in TVMCE are limited to a few scenes (stages, stadiums, and concerts) and specific types (broadcast or live stream).
We found that a multi-camera view recommendation model trained on TVMCE has issues generalizing to never-before-seen domains, and its accuracy drops significantly \red{(31.83 vs. 22.65)} when applied to other video scenes (e.g., living rooms) and/or types (e.g., movies).

Collecting data from the same domain would be the most direct approach to solving the issue.
Yet, such requires multiple synchronized cameras capturing the same event, not to mention the dedicated cinematography expertise required for labeling.

This paper proposes a methodology for generating pseudo-labeled multi-camera editing data from regular videos with shots, alleviating data scarcity.
With the proposed approach, sufficient data on a given domain (e.g., movie scenes in living rooms) could be obtained.
Our insights stem from two observations. (1) Many shot transitions within a regular video result from camera switches\footnote{\red{Other transitions such as video effect and trimming could be filtered by heuristics or treated as noises.}}.
Namely, videos are edited from their original multi-camera videos.
(2) In a multiple-camera system, cameras often remain stationary (extrinsics and intrinsics) and are usually responsible for shots of different scales.
Based on these two insights, we perform clustering on shots in a video to simulate different cameras and select the most visually similar shot from each camera as candidates alongside the ground truth to generate pseudo-labeled data.
A model trained on the proposed dataset enjoys a significant improvement in accuracy in the target domain (22.65 vs. 38.14\red{, cf. Table \ref{tab:domain}}).
\textbf{Our contributions} are summarized as follows:
\begin{itemize}
    \item We identify the poor domain generalizability of multi-camera view recommendation models. 
    \item We propose generating pseudo-labeled multi-camera editing datasets with regular videos to mitigate the lack of labeled data on an arbitrary domain.
    \item With the proposed pseudo-labeled multi-camera editing datasets, we achieve a 68\% relative improvement in the model's classification accuracy in the target domain. (\red{cf. Table \ref{tab:domain}}).
\end{itemize}
\section{Related Work}
\noindent Many works have studied multi-camera view recommendation \cite{wang2014context, smart_director,hu2023reinforcement,chen2019learning,biswas2024view,tvmce}.
\cite{wang2014context} designs two modules that collaboratively make switching decisions on a soccer game based on heuristics like object proximity and view duration.
\cite{smart_director} detects pre-defined events of interest in soccer games and leverages a scheduler to decide which view to broadcast.
These works design heuristics for a specific event or scene, limiting their applicability to other domains.
\cite{hu2023reinforcement} trains a reinforcement agent to predict video attributes for retrieving the most appropriate view.
Similar to our work, \cite{chen2019learning} leverages off-the-shelf videos as an auxiliary to augment the labeled dataset.
The works mentioned above, despite promising, require labeled multi-camera datasets in the target domain.
\cite{tvmce} puts up a larger-scale multi-camera editing dataset to foster the growth of the sphere. Yet, the dataset is still limited to particular video styles and cannot be generalized to other domains.
Our work differs from the previous works in that we propose a method to transform regular videos of an arbitrary domain into pseudo-labeled multi-camera editing datasets and \textit{altogether bypass the need for human-labeled datasets in the target domain}. 
\section{Framework}
\noindent\textbf{Problem Setup.}
This work aims to assist professionals in video editing by recommending camera tracks to switch to, considering the past shots that the professional has selected.
Note that we only focus on which track to switch to, but not whether to switch tracks because multi-camera view recommendation (1) is still nascent and (2) involves a certain amount of subjectivity.

The task is formulated as follows: $N$ temporally synchronized cameras capture the same scene from different angles, producing $N$ video tracks.
Given a portion of the past video that has been edited from time $T=s$ to $T=e$ of length $e-s$, the task is to decide which track to switch to at the coming time $T=t$, where $t-e$ could be variable.

\begin{figure*}
    \centering
\includegraphics[width=\linewidth]{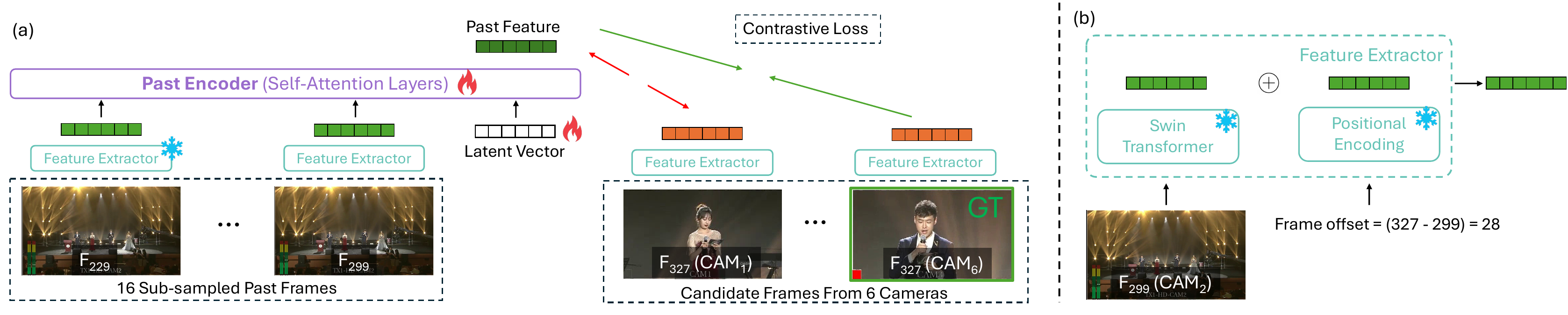}
    \caption{
    Model Architecture.
    (a) The past encoder encodes all past features to a single feature vector.
    Then, a contrastive loss is applied to maximize the cosine similarity between the past and ground-truth features.
    (b) The feature extractor encodes a frame by adding a positional embedding to the image feature.
    \raisebox{-\mydepth}{\includegraphics[height=\myheight]{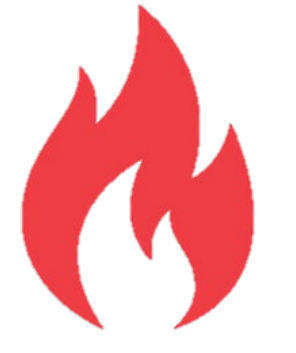}}: trainable, \raisebox{-\mydepth}{\includegraphics[height=\myheight]{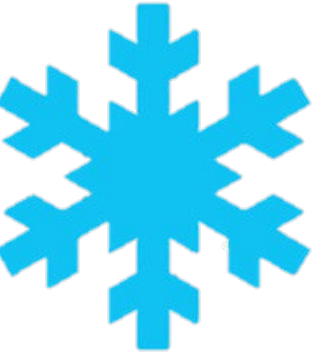}}: frozen. GT: groundtruth.
    The number F$_{\textit{N}}$ indicates the \textit{$N^{th}$} frame from the video.
    \textbf{[Best viewed in color.]}
    }
    \label{fig:arch}
\end{figure*}

As shown in Fig. \ref{fig:arch} (a), 16 sub-sampled past selected frames (one out of every five frames) are given as input to the model.
The output is which of the six cameras should be switched to $N_F$ frames away. Note that $N_F$ could be variable.

\vspace{0.5em}
\noindent\textbf{Model Architecture.}
As shown in Fig. \ref{fig:arch}(a), the model consists of two main modules: a \textit{feature extractor} encoding individual frames and their corresponding metadata into a single vector, and a \textit{past encoder} aggregating all past features into a learnable latent vector through layer(s) of self-attention.
This latent vector encapsulates holistic information from all the selected frames in the near past, such as visual cues and transitions between past frames.
In Fig. \ref{fig:arch}(b), the feature extractor has two inputs: the frame and its corresponding frame offset to the candidate frames.
A Swin Transformer \cite{swint} pre-trained on ImageNet \cite{imagenet} encodes the image into an image vector.
A positional embedding encodes how distant the input frame is from the candidate frames. For instance, in Fig. \ref{fig:arch}(b), the frame offset between the most recent past frame and the candidate frames is (327 - 299) = 28. We use sine and cosine functions of different frequencies as the positional embeddings \cite{transformer}.
We only train the past encoder and the latent vector.

\vspace{0.5em}
\noindent\textbf{Training Objectives.}
Following previous work on self-supervised feature learning \cite{moco, simclr}, we optimize the model by InfoNCE \cite{info_nce} and maximize the cosine similarity between the past feature and the ground-truth candidate feature while minimizing the cosine similarity between the past feature and all the other candidate features. 
\begin{figure}
    \centering
\includegraphics[width=\linewidth]{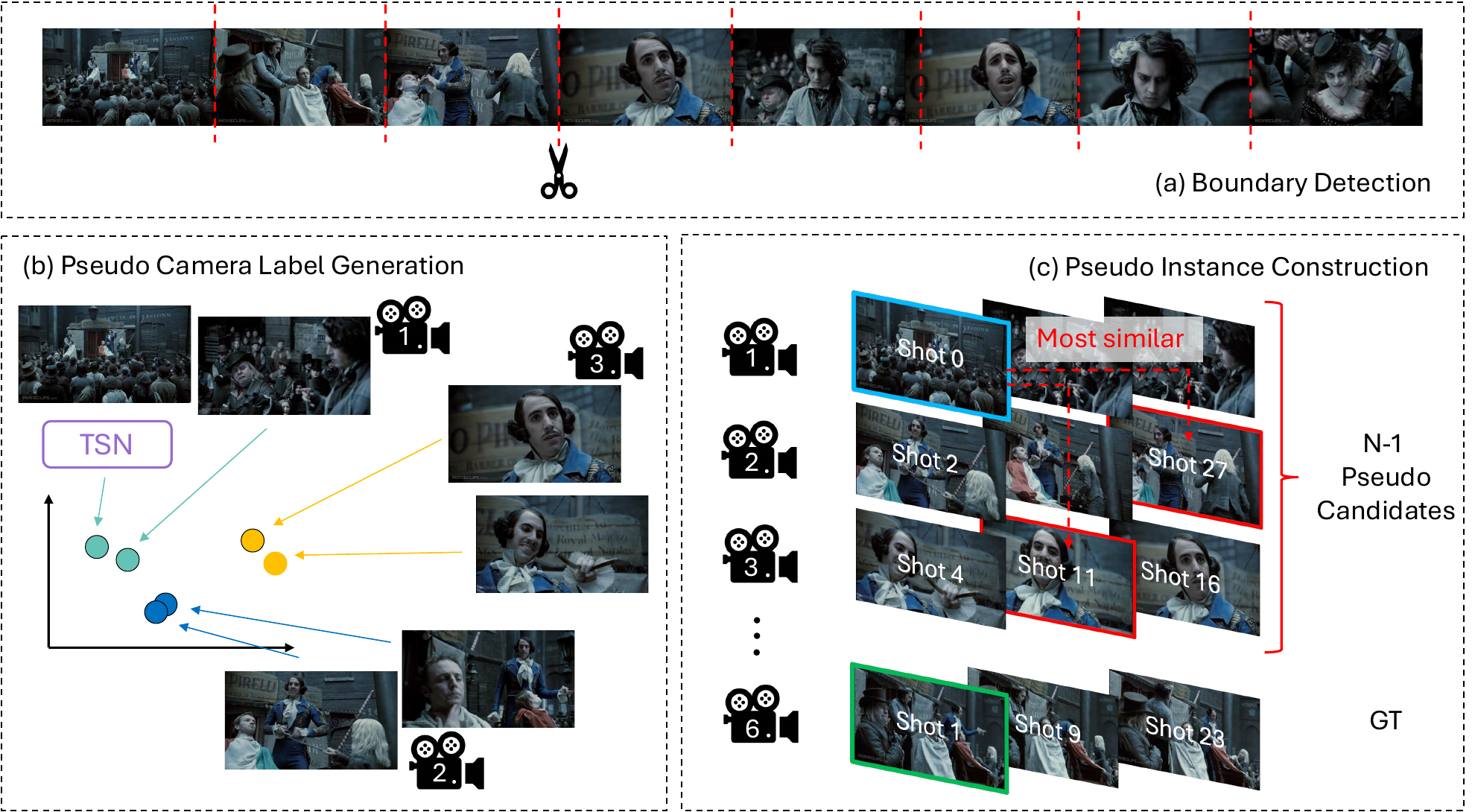}
    \caption{
        \textbf{Pseudo Dataset Generation Pipeline.} (a) Shots are detected in the input video, and
        (b) clustered into groups. Shots within the same cluster are regarded as from the same ``pseudo'' camera.
        (c) A shot is selected as an \textcolor{blue}{anchor}. The succeeding shot is the \green{ground truth}, while the most similar shot amongst each of the other N-1 pseudo cameras is chosen as a \textcolor{red}{candidate}. \textbf{[Best viewed in color with zoom-in.]}
    }
    \label{fig:pseudo_dset}
    % \vspace{-1.3em}
\end{figure}

\section{Pseudo Multi-Camera Editing Dataset}
\noindent As mentioned in Sec. \ref{sec:intro}, to deal with the scarcity of labeled multi-camera editing data, we propose transforming regular videos in the target domain to pseudo-labeled multi-camera editing data.
This section demonstrates the detailed procedure of transforming regular videos into pseudo-labeled multi-camera editing datasets.
Fig. \ref{fig:pseudo_dset} illustrates creating a pseudo dataset. 
First, we detect shots in a video and create pseudo-cameras by clustering the shots. Finally, we select candidates from each pseudo-camera (cluster) to construct an instance. 

\vspace{0.5em}
\noindent \textbf{Camera Switch.}
% \label{subsec:shot_detect}
Our insight stems from the observation that 
% \textit{shot transitions result from camera switches}.
when a professional is editing multi-camera videos, a shot boundary is when they decide to switch from one camera to another. 
Though we cannot access the original multi-camera videos, the final video carries partial supervision from which the model could learn. 
We first perform shot detection on a given video and obtain $N_S$ shots.

\vspace{0.5em}
\noindent \textbf{Pseudo Camera Label Generation.}
% Since we cannot access the original multi-track videos, we need a way to generate pseudo-multi-tracks.
We made two observations. First, in a multi-camera system, each camera tends to remain stationary and keeps its perspective.
Second, cameras are usually responsible for shots of different scales.
% 
% For example, one camera is responsible for long shots, while another could be responsible for extreme close-ups.
%  
Utilizing the insight that \textit{shot scale of each camera tends to stay unchanged}, we train a  Temporal Segment Network (TSN) \cite{tsn} for shot type classification on MovieShots \cite{movieshot} to extract features for each shot.
Then, we obtain six clusters by running K-Means on the features. Each cluster is treated as a pseudo-camera.

\vspace{0.5em}
\noindent \textbf{Pseudo Instance Construction.}
% \label{subsec:shot_sim}
After assigning a pseudo-camera label to every video shot, we select the most visually similar shot from each pseudo-camera as the candidates alongside the ground truth, as in Fig. \ref{fig:pseudo_dset}.
We use a ResNet50 \cite{resnet} pre-trained on ImageNet \cite{imagenet} to extract image features for the first and last frames in each shot and normalize the features to unit length.
% 
% Since we focus on the transition between shots, we only consider image features from the first and the last frames when calculating shot similarity.
% 
Then, we compute the cosine similarity of the previous shot's last-frame feature to the next shot's first-frame feature as the visual similarity between the two shots.
\section{Experiment}

\subsection{Setup}
\noindent \textbf{Datasets.}
% \label{subsec:datasets}
We leverage four datasets: \textbf{TVMCE} \cite{tvmce} and three pseudo datasets created from \textbf{ClipShots} \cite{clipshot}, \textbf{Condensed Movies} \cite{condensed_movie}, and \textbf{Sitcoms} episodes.
\textbf{TVMCE} consists of 6 synchronized camera tracks, totaling 88 hours of recorded videos.
The professionals edit the multi-camera videos into a single-track video, providing 5133 ground-truth camera transition labels.
Following \cite{tvmce}, 4042 and 1091 transitions are used for training and testing, respectively.
\textbf{ClipShots} \cite{clipshot} is a dataset for shot boundary detection.
It contains videos from over 20 categories, ranging widely from movie spotlights to phone videos. 
After removing shots with gradual transitions, e.g., fade, we leverage the remaining shots for pseudo-dataset generation. In total, there are 118658 camera transitions.
\textbf{Condensed Movies} \cite{condensed_movie} is proposed to understand the narrative structure of movies and contains key scenes from over 3K videos, providing 152893 camera transitions.
Shots from one episode of each of the four \textbf{Sitcoms}: Friends, How I Met Your Mother, The Big Bang Theory, and Two and a Half Men are used for pseudo dataset generation.

\vspace{0.5em}
\noindent \textbf{Implementation Details.} 
All models are trained for ten epochs, with a learning rate of 1e-5 and a batch size of 2. The mean and standard deviation of results from three seeds are reported for each setup.
TransNet V2 \cite{trans_net} is used for shot detection. TSN pre-trained on MovieShots \cite{movieshot} for 60 epochs, reaching a top-1 shot scale classification accuracy of 90.08\%, is used for feature extraction. Specifically, features from the penultimate layer are used for shot clustering.
We remove videos with fewer than ten shot transitions, as transitions in these videos are likely from video effects but not actual camera transitions. We also discard shots with gradual transitions. 
For evaluation, \textbf{Classification accuracy} of the camera being switched to is used.

\subsection{Result}
\noindent \textbf{Baseline Comparison.} 
A few works have been proposed for multi-camera view recommendation \cite{chen2019learning, hu2023reinforcement, smart_director}. Yet, neither their codes nor their datasets are publicly available.
We re-implement the Temporal and Contextual Transformer (TC Transformer) in \cite{tvmce} as the primary baseline we compare to.
It achieves state-of-the-art (SoTA) classification accuracy in multi-camera view recommendation on TVMCE dataset.
Note that we remove the training data in the middle of a shot (without camera switches), which improves the accuracy (22.48 in \cite{tvmce}) of TC Transformer to 28.77.
As shown in Table \ref{tab:baselines}, the proposed framework outperforms TC Transformer by 11\% (\red{31.83} vs. 28.77).
However, both models suffer when applied to a never-before-seen domain.

\begin{table}
    \normalsize
    \centering
    \setlength{\tabcolsep}{5pt}
    \begin{tabular}{c|c|c}
        \toprule
        Method$\downarrow$ / Test Set$\rightarrow$ & TVMCE (ID) & Sitcoms (OOD) \\
        \midrule
        Random & 16.67 & 16.67 \\
        TC Transformer & 28.77±0.86 & 14.04±1.59 \\
        Ours & \textbf{\red{31.83±0.83}} & \textbf{22.65±2.43} \\
        \bottomrule
    \end{tabular} \\
    \caption{
    \textbf{Comparison to Baseline.}
    Multi-camera view recommendation models generalize poorly to a never-before-seen domain.
    ID: in-Domain, OOD: out-of-Domain.
    }
    \label{tab:baselines}
\end{table}

\vspace{0.5em}
\noindent \textbf{Generalizability \& Pseudo Dataset Efficacy.} 
We investigate domain differences from two perspectives: (1) \textit{video scene}, and (2) \textit{video type}. Video scene is where the videos are filmed, e.g., on stage or in a livingroom. Video type is how the contents are presented, e.g., a live performance or a movie.

\red{First, we fix video type and examine the impact of video scene.}
Concretely, we split the TVMCE training set into two roughly equal-sized subsets based on the video scenes.
One subset, \textbf{TVMCE\textunderscore stage}, similar to the test set, consists of videos of stages and concert halls. The other subset, \textbf{TVMCE\textunderscore sport}, contains sports videos.
In Table \ref{tab:tvmce_domain}, the model trained on the dissimilar scenes achieves significantly lower accuracy than one trained on the same scenes as the test set.

Next, we investigate the impact of both video scenes and video types.
In this experiment, all models are evaluated on \textbf{Sitcoms}, consisting of videos that differ in both scenes and types from those in TVMCE.
Specifically, video scenes in TVMCE are mostly stages and sports, whereas video scenes in Sitcoms are kitchens and living rooms. Also, TVMCE contains recorded live performances, whereas Sitcoms contains movies.
Three models are evaluated, each trained on a dataset with different degrees of domain difference to Sitcoms. (1) TVMCE that differs in both video type and scenes, (2) ClipShots differs only in video type, and (3) Condensed Movies that covers the same video type and scenes.
Two observations can be made in Table \ref{tab:domain}.
(1) Pseudo datasets bridge the data scarcity and improve the accuracy in the target domain.
(2) Video type and scene can compound to impact performance. TVMCE, different in both video type and scenes from Sitcoms, performs the worst.
On the contrary, Condensed Movies with the same video type and scenes as Sitcoms achieves the best performance.

Better accuracy of the model trained on pseudo multi-camera editing datasets --- ClipShots in Table \ref{tab:tvmce_domain}, and ClipShots and Condensed Movies in Table \ref{tab:domain} demonstrate their efficacy in improving model performance in the target domain without labeling by professionals.

\red{
Note that previous work also leverages pseudo labels for evaluation \cite{hu2023reinforcement}; still, an important question is whether pseudo-labeled datasets reflect the performance of human-labeled ones.
We treat the TVMCE test set as unlabeled, construct a pseudo-TVMCE test set, and evaluate the model trained with sports scenes on this set.
The accuracy is similar to the TVMCE test set (23.71±0.12 vs. 25.56±0.87).
}

\begin{table}
    \centering
    \setlength{\tabcolsep}{10pt}
    {\small TVMCE test set --- video scenes: \green{stage, concert hall}} \\
    \normalsize
    \begin{tabular}{c|c|c}
        \toprule
        Train Set & Video Scene & Accuracy \\
        \midrule
        TVMCE\textunderscore stage & \green{stage, concert hall} & 31.83±0.83 \\
        TVMCE\textunderscore sport & \textcolor{red}{sports} & 25.56±0.87 \\
        \midrule
        \midrule
        ClipShots \textbf{(Pseudo)} & \green{broad\textsuperscript{1}} & 26.81±0.22 \\
        \bottomrule
    \end{tabular}  \\
    {\footnotesize \textsuperscript{1}consists of videos from a broad scene coverage.}
    \caption{
    % \small
    \textbf{Impact of Video Scene.} The model trained in different scenes to the test set achieves lower accuracy.
    \green{Green}, and \textcolor{red}{red} means same and different.
    \textbf{[Best viewed in color.]}
    }
    \label{tab:tvmce_domain}
\end{table}

\vspace{0.5em}
\noindent \textbf{Pseudo Dataset Construction Methodology.}
\red{Three methodologies are investigated: (1) selecting the most visually similar shot from each cluster, (2) selecting one random shot from each cluster, and (3) selecting the five most similar shots \textit{without clustering}.
On the TVMCE test set, the three models achieve an accuracy of 26.81±0.22, 7.16±0.40, and 26.02±0.49, respectively.}
The discrepancy between (1) and (2) shows the importance of selecting the shots that are most visually similar to the candidates.
We conjecture that the ground-truth succeeding shots tend to have a certain amount of overlap in viewpoint with their predecessors, and randomly selected shots with little visual overlap to the previous shot could be quickly ruled out by the model, leading to ineffective training.
The difference between (1) and (3) shows that while clustering cannot recover the actual cameras, it can be used to obtain hard negatives, thus facilitating model training.

\begin{table}
    \normalsize
    \centering
    {\small Test set --- video type: \green{movie}, video scenes: \green{living room, kitchen}} \\
    \begin{tabular}{c|c|c|c}
        \toprule
        Train Set & Video Type & Video Scene & Accuracy \\
        \midrule
        TVMCE & \textcolor{red}{live} & \textcolor{red}{stages, sports} & 22.65±2.43 \\
        \midrule
        ClipShots \textbf{(Pseudo)} & \orange{mixed\textsuperscript{1}} & \green{broad\textsuperscript{2}} & 27.61±1.20 \\
        Condensed \textbf{(Pseudo)} & \green{movie} & \green{broad\textsuperscript{2}} & \textbf{38.14±1.50} \\
        \bottomrule
    \end{tabular} \\
    \footnotesize{\textsuperscript{1}consists of videos of movies and live performance. 
    
    \textsuperscript{2}consists of videos from a broad scene coverage.
    }
    \caption{
    \textbf{Impact of Video Scene and Type (Sitcoms).}
    A more significant domain difference (video scenes + video types) severely impacts accuracy.
    Pseudo datasets from a broad range of videos could cover the target video scenes and types, achieving better accuracy.
    \green{Green}, \orange{orange}, and \textcolor{red}{red} means same, covered, and different. 
    \textbf{[Best viewed in color.]}
    }
    \label{tab:domain}
\end{table}
\section{Discussion} \label{sec:discussion}
\noindent \textbf{Application Scenarios.}
Based on the experiment results, we divide multi-camera view recommendation into three scenarios.
(1) In a controlled environment where decisions made by professionals for similar events are accessible.
For example, in soccer games, the camera transition patterns are relatively straightforward, and multi-camera videos are available alongside the final broadcast. 
In this scenario, supervised learning on the professional-labeled datasets would suffice.
(2) If the model were to be applied to domains where decisions made previously by experts are unavailable, we would suggest generating pseudo datasets on the target domains with off-the-shelf regular videos for training.
(3) If one does not know the explicit target domains and wants to apply the model in the wild, the best strategy would be to collect videos from extensive and different domains and train the model on pseudo datasets transformed from these videos. This would increase the probability of the pseudo datasets covering the target domains.
We note that the proposed pseudo multi-camera editing dataset \textit{does not solve the \red{innate} domain gap issue} once and for all but mitigates it through the availability of regular videos in an arbitrary domain.
\section{Conclusion}
In this paper, we first show that multi-camera view recommendation models struggle to generalize to never-before-seen domains. We then analyze two aspects that could compound and intensify domain mismatch: video scenes and video types. 
% 
% We attempt to resolve the issue by first improving the model accuracy in the training domain by introducing two components: frame offset embedding and camera embedding.
% 
% Despite our success in improving the baseline accuracy by 29\%, we show that simply improving a model in the training domain does not enhance its generalizability.
% 
We propose to leverage regular videos in the target domain to generate pseudo multi-camera editing datasets. We also design a learning framework that optimizes the model parameters with contrastive loss to bring close the current and the succeeding ground-truth shots. 
Experiments demonstrate the efficacy of the proposed pseudo multi-camera editing dataset in improving the model's accuracy in the target domain.

Future work involves designing a better training objective to capture the underlying cinematography expertise, which enables professionals to work across video scenes and types.
Another potential direction is to incorporate explicit rules of thumb and conventions in cinematography, e.g., the gradual change from wider to narrower shots and the rule of thirds. 
% 
% Either transforming these best practices into objectives or framing them as modules in a mixture of experts is worth investigating.

\section{Acknowledgement}
\noindent This work was funded by National Science Foundation grants NSF CNS 19-00875, NSF CCF 22-17144, NSF CNS 21-06592. Any results and opinions are our own and do not represent views of National Science Foundation.

\bibliographystyle{plain}
\bibliography{reference}

\end{document}